%% file: main.tex
\documentclass{article}

\usepackage{template/iclr2026/iclr2026_conference,times}
\input{math_commands.tex}

\input{preamble.tex}

\usepackage{hyperref}
\usepackage{url}
\hypersetup{hidelinks}

\ifdefined\iclrsubmissioncopy\else
\iclrfinalcopy
\fi

\title{Decoder-Preserving Sparse Autoencoders:\\Which Readouts Survive Sparse Compression?}
\author{Aniket Deshpande \\
  Department of Physics, University of Illinois Urbana-Champaign \\
  \texttt{aniket4@illinois.edu}
}

\begin{document}
\raggedbottom

\maketitle
\ifdefined\iclrsubmissioncopy\else
\lhead{Preprint}
\fi

\begin{abstract}
{\color{black}
Sparse autoencoders (SAEs) compress model activations into sparse codes, but equal reconstruction error and sparsity can preserve different linearly decodable signals. We formalize this ambiguity as a matrix-valued distortion between optimal ridge-prediction operators and train decoder-preserving SAEs by combining this distortion with reconstruction loss. In a rank relaxation, an isotropic task prior saturates per-mode omission costs without changing PCA's ordering, whereas a structured prior can change which modes are retained. A controlled sparse experiment shows that a declared prior protects held-out combinations from its task subspace. On GPT-2 small block 8, DPSAE reduces held-out decoder distortion by 10.6--11.4\% across three paired runs while matching reconstruction NMSE. The same checkpoints pass an average natural-text output-KL noninferiority test, but one matched Pythia pair shows no improvement in probes restricted to a few sparse features. These results show that reconstruction quality does not determine which refitted linear readouts survive sparse compression, and that readout preservation is distinct from learning cleaner benchmark concepts or preserving every frozen-model behavior.
\begin{center}
\href{https://github.com/aniket-desh/decoder-preserving-sae}
     {\texttt{github.com/aniket-desh/decoder-preserving-sae}}
\end{center}
}

\end{abstract}

\input{sections/introduction}
\input{sections/background}
\input{sections/method}
\input{sections/experiments}
\input{sections/results}
\input{sections/related_work}
\input{sections/conclusion}

\begingroup
\setlength{\bibsep}{2pt plus 0.5pt minus 0.5pt}
\bibliography{references}
\bibliographystyle{template/iclr2026/iclr2026_conference}
\endgroup

\newpage
\appendix
\input{appendices/experimental_details}
\input{appendices/theory_proofs}

\end{document}

%% file: math_commands.tex
\newcommand{\R}{\mathbb{R}}
\newcommand{\Xhat}{\widehat{X}}

\newcommand{\tr}{\operatorname{tr}}
\newcommand{\E}{\mathbb{E}}

%% file: preamble.tex
\usepackage{amsmath,amssymb,amsthm,mathtools}
\usepackage{booktabs}
\usepackage{float}
\usepackage{graphicx}
\usepackage{microtype}
\usepackage{xcolor}

\newcommand{\rev}[1]{#1}

\newtheorem{theorem}{Theorem}
\newtheorem{proposition}[theorem]{Proposition}
\newtheorem{corollary}[theorem]{Corollary}

%% file: sections/introduction.tex
\section{Introduction}
\label{sec:introduction}

Sparse autoencoders (SAEs) compress model activations into sparse feature dictionaries for analysis and intervention \citep{karvonen2025saebench}. When exact reconstruction is unavailable at a fixed sparsity budget, error must fall somewhere in activation space. MSE assigns capacity by squared displacement in that space. Many analyses instead ask whether linearly recoverable variables remain recoverable after compression. Two reconstructions produced from equally sparse codes can therefore match exactly in MSE while allocating their errors to different decoding tasks.

We ask whether an SAE can preserve a family of linear readouts more effectively without increasing reconstruction MSE. Harvey et al. interpret several representation-similarity measures through agreement across optimal linear readouts \citep{harvey2024representational}. For a fixed batch, $K_X$ maps any target vector to the predictions of the optimal regularized linear probe fitted to $X$. Comparing it with the corresponding operator for a reconstruction measures the taskwise pattern of ridge-prediction distortion. We use this prediction-operator geometry as a sparse-compression objective and analyze the taskwise underdetermination left by MSE and sparsity.

Readout fidelity is therefore task-indexed and matrix-valued. A task second moment weights the decoding tasks that contribute to its scalar summary. We combine decoder disagreement with MSE and fixed sparsity, and call the resulting model a decoder-preserving sparse autoencoder (DPSAE). MSE anchors the reconstruction in the coordinate system expected by the language model. The decoder term biases the sparse bottleneck toward readouts weighted by the task prior. An isotropic prior weights target directions equally, whereas a structured prior can emphasize a declared task subspace.

\begin{figure}[!htbp]
  \centering
  \includegraphics{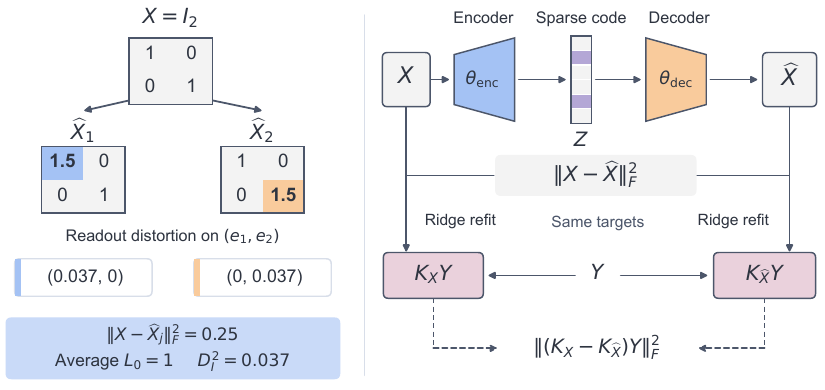}

  \caption{\rev{\textbf{Equal reconstruction error leaves readout fidelity underdetermined.} Left: with $W=I_2$ and $Z_j=\widehat X_j$, the two reconstructions have the same squared Frobenius error $\lVert X-\widehat X_j\rVert_F^2=.25$, average sparsity $L_0(Z_j)=1$, and isotropic decoder distance $D_I^2=.037$. At $\lambda=.5$, their squared ridge-prediction distortions on targets $(e_1,e_2)$ are $(.037,0)$ and $(0,.037)$, respectively. Right: DPSAE combines reconstruction error with disagreement between separate ridge refits on $X$ and $\widehat X$ using the same target matrix $Y$.}}
  \label{fig:conceptual-theory}
\end{figure}

\rev{Figure~\ref{fig:conceptual-theory} makes this freedom explicit and summarizes the training objective.}

\begin{samepage}
We make three contributions:
\begin{itemize}
  \item \parbox[t]{\linewidth}{\textbf{Taskwise underdetermination and its rank-relaxed limit.} We prove that equal MSE and code $L_0$ do not determine ridge-readout fidelity. Under a pure rank bottleneck, isotropic decoder preservation keeps PCA's singular-direction ordering but replaces variance-proportional omission cost with a ridge-saturated cost. Any qualitative departure from this rank-relaxed ordering must involve structure omitted by the relaxation, such as sparse support allocation.}
  \item \parbox[t]{\linewidth}{\textbf{Task-selective preservation under declared priors.} A task second moment scalarizes the distortion operator. In the commuting rank relaxation, a structured task prior retains modes according to a closed-form weighted score. In a controlled sparse generator, the prior reduces median distortion on unseen combinations from its declared subspace by 25.3\% relative to paired MSE, with no observed NMSE penalty.}
  \item \parbox[t]{\linewidth}{\color{black}\textbf{Language-model evidence.} On GPT-2 small block 8, DPSAE reduces exact finite-group decoder distortion by 10.6--11.4\% across three paired runs at matched reconstruction quality. The same pairs pass average natural-text output-KL noninferiority but perform worse on IOI; one matched Pythia pair also shows no gain in sparse concept probes.\color{black}}
\end{itemize}
\end{samepage}

\rev{These claims concern regularized readouts refitted after compression. The frozen-network and concept-probe results show why they should not be read as evidence for uniform downstream fidelity, cleaner sparse concepts, or cross-model generality.}

%% file: sections/background.tex
\section{Background}
\label{sec:background}

\subsection{Sparse autoencoders}

Sparse autoencoders learn a dictionary for model activations. An encoder maps an activation $h\in\R^d$ to a sparse code $z\in\R^m$, and a decoder reconstructs $\hat h$ as a weighted combination of its dictionary vectors. Language-model applications use the active coordinates of $z$ as candidate features for analysis and intervention \citep{huben2024sparse,bricken2023monosemanticity}.

Standard SAE training balances reconstruction error against sparsity. An $\ell_1$ penalty encourages small codes but controls sparsity only indirectly. TopK variants instead retain a fixed number of code entries, which makes the reconstruction--sparsity tradeoff easier to compare across models \citep{gao2025scaling}. In either case, squared reconstruction error assigns capacity according to Euclidean displacement in activation space. It does not directly measure whether a target remains predictable from the reconstruction.

\subsection{Linear decodability as representation fidelity}

A linear decoding task fits a readout from a representation matrix $X\in\R^{n\times d}$, with one example per row, to targets defined on the same examples. Ridge regression makes this readout well posed by penalizing its weight norm. Each representation then induces a prediction operator $K_X$ that maps any target vector to the predictions of its optimal ridge readout. Comparing $K_X$ with the operator induced by a reconstruction asks whether the same family of targets remains accessible after refitting.

This prediction-based view also underlies existing representation distances. GULP compares representations through regularized linear prediction tasks and controls differences in their predictive performance \citep{boixadsera2022gulp}. Harvey et al. show that CKA, CCA, and related similarity measures can likewise be interpreted as average alignment between optimal linear readouts over distributions of decoding tasks \citep{harvey2024representational}. These are statements about refitted decoders. They do not imply that a frozen downstream network will process a reconstructed activation in the same way, because its weights are not refitted.

%% file: sections/method.tex
\section{Readout Fidelity Under Compression}
\label{sec:method}

\subsection{Readout distortion is matrix-valued}

Let $X\in\R^{n\times d}$ contain one activation per row. For targets $y\in\R^n$, the prediction of the optimal ridge readout from $X$ is
\begin{equation}
  \hat y_X = K_X y,
  \qquad
  K_X = X(X^\top X + n\lambda I)^{-1}X^\top,
  \label{eq:ridge-hat}
\end{equation}
where $\lambda>0$ is the penalty in the average ridge objective $n^{-1}\lVert Xw-y\rVert_2^2+\lambda\lVert w\rVert_2^2$. Given a reconstruction $\Xhat$, define the signed prediction-operator error and its taskwise distortion operator as
\begin{equation}
  \Delta_X(\Xhat)=K_X-K_{\Xhat},
  \qquad
  \mathcal A_X(\Xhat)=\Delta_X(\Xhat)^\top\Delta_X(\Xhat).
  \label{eq:distortion-operator}
\end{equation}
The distortion of a target $y$ is the quadratic form
\begin{equation}
  d_{\Xhat}(y)
  =y^\top\mathcal A_X(\Xhat)y
  =\lVert K_Xy-K_{\Xhat}y\rVert_2^2.
  \label{eq:task-distortion}
\end{equation}
For a task distribution with second moment $\Sigma_y=\E[yy^\top]\succeq0$, its exact decoder distance and source prediction energy are
\begin{equation}
\begin{aligned}
  D_{\Sigma_y}^2(X,\Xhat)
  &=\tr\!\left(\Sigma_y\mathcal A_X(\Xhat)\right),
  \qquad
  S_{\Sigma_y}(X)
  =\tr\!\left(\Sigma_y K_X^\top K_X\right).
\end{aligned}
\label{eq:decoder-distance}
\end{equation}
When $S_{\Sigma_y}(X)>0$, the exact relative distortion is $R_{\Sigma_y}(X,\Xhat)=D_{\Sigma_y}^2(X,\Xhat)/S_{\Sigma_y}(X)$. Readout distortion is therefore matrix-valued before a task second moment scalarizes it. The isotropic choice $\Sigma_y=I$ gives $D_I^2=\lVert K_X-K_{\Xhat}\rVert_F^2$.

\subsection{MSE and sparsity do not determine taskwise fidelity}

Reserve $Z\in\R^{n\times p}$ for the sparse code and define average per-example sparsity as
\begin{equation}
  L_0(Z)=\frac1n\sum_{i=1}^n\lVert Z_{i,:}\rVert_0.
  \label{eq:average-l0}
\end{equation}
The freedom left by MSE and $L_0$ is exact, even in two dimensions.

\begin{proposition}[Equal-error taskwise separation]
\label{prop:equal-error-separation}
For every $\lambda>0$, there exist a source $X\in\R^{2\times2}$, nonnegative codes $Z_1,Z_2\in\R_+^{2\times2}$, and a common linear decoder $W$ with reconstructions $\Xhat_j=Z_jW$ such that
\[
  L_0(Z_1)=L_0(Z_2)=1,
  \qquad
  \lVert X-\Xhat_1\rVert_F^2=\lVert X-\Xhat_2\rVert_F^2,
\]
and $D_I^2(X,\Xhat_1)=D_I^2(X,\Xhat_2)$. Nevertheless, $\mathcal A_X(\Xhat_1)$ and $\mathcal A_X(\Xhat_2)$ are incomparable in the Loewner order. Each reconstruction has lower distortion than the other on some unit target.
\end{proposition}

The construction changes one diagonal coordinate in each reconstruction. It gives the same MSE, code sparsity, and isotropic average distortion while exchanging which coordinate readout is preserved. Appendix~\ref{app:theory-proofs} gives the construction and proof.

For two trained reconstructions $\Xhat_M$ and $\Xhat_D$, define their DPSAE-advantage operator
\begin{equation}
  Q=\mathcal A_X(\Xhat_M)-\mathcal A_X(\Xhat_D).
  \label{eq:advantage-operator}
\end{equation}

\begin{proposition}[Taskwise advantage characterization]
\label{prop:advantage-spectrum}
For every target $y$, $d_{\Xhat_M}(y)-d_{\Xhat_D}(y)=y^\top Qy$. Over unit targets, the smallest and largest advantages are $\lambda_{\min}(Q)$ and $\lambda_{\max}(Q)$. DPSAE has no greater distortion on every target exactly when $Q\succeq0$, and it has lower $\Sigma_y$-weighted distortion exactly when $\tr(\Sigma_yQ)>0$. If $Q$ is indefinite, rank-one task priors can prefer either reconstruction.
\end{proposition}

This proposition is a diagnostic rather than the explanation for why training changes a reconstruction. Its trace measures isotropic average advantage, while its eigenvalues distinguish average improvement from uniform taskwise dominance.

\subsection{The rank relaxation isolates what isotropy can change}

The simplest relaxation replaces an SAE by any reconstruction of rank at most $r$. This removes elementwise sparsity and nonlinear support selection while retaining the prediction-operator objective.

\begin{theorem}[Isotropic rank relaxation]
\label{thm:isotropic-rank-relaxation}
Let $X=U\operatorname{diag}(\sigma_1,\ldots,\sigma_s)V^\top$ be a compact SVD with $\sigma_1\geq\cdots\geq\sigma_s>0$, and define
\[
  q_i=\frac{\sigma_i^2}{\sigma_i^2+n\lambda}.
\]
For $0\leq r\leq s$,
\begin{equation}
  \min_{\operatorname{rank}(\Xhat)\leq r}
  \lVert K_X-K_{\Xhat}\rVert_F^2
  =\sum_{i>r}q_i^2.
  \label{eq:isotropic-rank-optimum}
\end{equation}
The truncated SVD $X_r$ attains the minimum.
\end{theorem}

Rank-$r$ MSE and isotropic decoder preservation use the same mode ordering but assign different costs to each omitted mode:
\begin{equation}
  c_i^{\mathrm{MSE}}=\sigma_i^2,
  \qquad
  c_i^{\mathrm{dec}}=
  \left(\frac{\sigma_i^2}{\sigma_i^2+n\lambda}\right)^2.
  \label{eq:mode-omission-costs}
\end{equation}
The decoder cost saturates at one once a direction is strongly decodable. Since it remains monotone in $\sigma_i$, both objectives retain the same family of optimal top left singular subspaces under a pure rank bottleneck, up to choices within singular-value ties. Isotropic decoder preservation cannot reorder singular directions in this relaxation. Any qualitative departure from this rank-relaxed ordering must involve ingredients omitted by the relaxation, including sparse support, overcompleteness, nonorthogonal features, unequal firing rates, nonlinear selection, minibatch estimation, or optimization.

\begin{figure}[!htbp]
  \centering
  \includegraphics{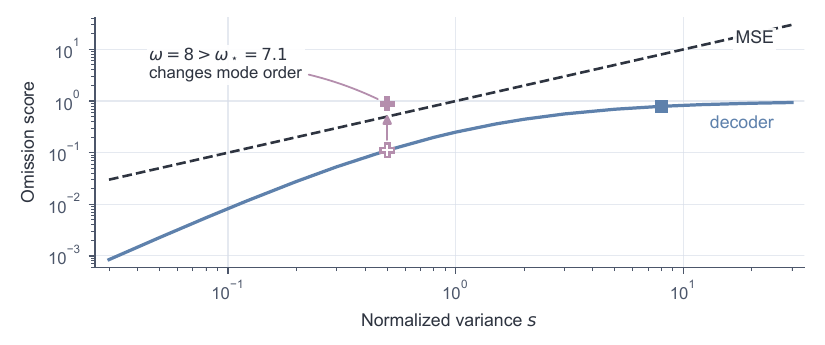}
  \caption{\rev{\textbf{Isotropic decoder omission cost saturates without changing the rank-relaxed mode order.} MSE omission cost grows with normalized variance $s$, whereas the isotropic decoder cost approaches one. Both remain monotone in $s$; the purple markers illustrate how a sufficiently strong structured task weight can cross the mode-selection threshold.}}
  \label{fig:spectral-mechanism}
\end{figure}

\subsection{A structured task prior can change the retained modes}

The monotone ordering above is specific to an isotropic task distribution. A structured prior can change it when the task and representation geometries align.

\begin{corollary}[Structured-prior selection]
\label{cor:structured-prior-selection}
Suppose $\Sigma_y\succeq0$ commutes with $XX^\top$. Choose a simultaneous eigenbasis in which
\[
  XX^\top=U\operatorname{diag}(\sigma_i^2)U^\top,
  \qquad
  \Sigma_y=U\operatorname{diag}(\omega_i)U^\top,
\]
and let $q_i=\sigma_i^2/(\sigma_i^2+n\lambda)$, including $q_i=0$ on the nullspace of $X$. For $0\leq r\leq\operatorname{rank}(X)$, if $S_r$ indexes the $r$ largest scores $\omega_iq_i^2$, then
\begin{equation}
  \min_{\operatorname{rank}(\Xhat)\leq r}D_{\Sigma_y}^2(X,\Xhat)
  =\sum_{i\notin S_r}\omega_iq_i^2.
  \label{eq:structured-rank-optimum}
\end{equation}
An optimizer retains the source modes in $S_r$ at their original singular values.
\end{corollary}

A protected direction $j$ overtakes a nuisance direction $i$ exactly when
\begin{equation}
  \omega_j
  \left(\frac{\sigma_j^2}{\sigma_j^2+n\lambda}\right)^2
  >
  \omega_i
  \left(\frac{\sigma_i^2}{\sigma_i^2+n\lambda}\right)^2.
  \label{eq:structured-crossover}
\end{equation}
The exact two-direction construction tests this crossover directly. The sparse generator uses a concatenated stochastic task prior whose expected covariance need not commute with the sample Gram matrix, so the same threshold is only a scale reference there, not a predicted sparse transition. Without the commutation assumption, an optimal weighted rank approximation can mix source modes, and the scalar ranking in Equation~\ref{eq:structured-crossover} need not hold.

\subsection{Why the objective remains hybrid}

Decoder distance identifies a larger equivalence class than activation reconstruction.

\begin{proposition}[Zero decoder distance]
\label{prop:zero-distance}
For fixed $\lambda>0$,
\begin{equation}
  K_X=K_{\Xhat}
  \quad\Longleftrightarrow\quad
  XX^\top=\Xhat\Xhat^\top.
  \label{eq:zero-distance}
\end{equation}
If $\Sigma_y\succ0$, these conditions are also equivalent to $D_{\Sigma_y}^2(X,\Xhat)=0$. When $X$ and $\Xhat$ have the same feature dimension, equal row Gram matrices imply $\Xhat=XQ$ for some orthogonal $Q$.
\end{proposition}

Decoder distance can therefore be zero after a feature-coordinate rotation that would change the input seen by the frozen downstream network. MSE anchors a member of this rotation-equivalence class to the original coordinates. The task prior still supplies a finite-group guarantee: for $y\in\operatorname{range}(\Sigma_y)$,
\begin{equation}
  \lVert\Delta_X(\Xhat)y\rVert_2^2
  \leq
  D_{\Sigma_y}^2(X,\Xhat)\,y^\top\Sigma_y^\dagger y.
  \label{eq:task-ellipsoid-bound}
\end{equation}
This bound controls the protected task ellipsoid on the observed group. It is not a generalization guarantee for new activation samples.

\subsection{\textcolor{black}{A scalable decoder-preservation objective}}

{\color{black}
For one geometry group, let $Y_m=m^{-1/2}[y_1,\ldots,y_m]$ contain task samples with $\E[y_jy_j^\top]=\Sigma_y$. We estimate relative decoder distortion as
\begin{equation}
  \widehat R_m(X,\Xhat;Y_m)
  =\frac{\lVert\Delta_X(\Xhat)Y_m\rVert_F^2}
  {\max\{\lVert K_XY_m\rVert_F^2,\epsilon\}}.
  \label{eq:sampled-relative-loss}
\end{equation}
The language-model experiments use fixed-radius Gaussian directions, while the controlled experiments use Rademacher probes and add scaled protected-subspace targets for the structured prior. Language-model training aggregates sampled numerators and denominators across geometry groups before taking their ratio. Identity targets compute the isotropic trace exactly on a fixed finite group. Appendix~\ref{app:estimator-details} defines the corpus aggregations and records the finite-probe, scaling, and denominator-clamp audits.

For encoder $E_\theta$, decoder $D_\theta$, sparse code $Z=E_\theta(X)$, and reconstruction $\Xhat=D_\theta(Z)$, the hybrid objective is
\begin{equation}
  \mathcal L(X)
  =\alpha\frac{\lVert X-\Xhat\rVert_F^2}{\lVert X\rVert_F^2}
  +\gamma\widehat R_{\mathcal G,m}(X,\Xhat)
  +\mathcal L_{\mathrm{sparse}}(Z),
  \label{eq:hybrid-loss}
\end{equation}
where $\widehat R_{\mathcal G,m}$ is the aggregation used by the corresponding experiment. The Euclidean term anchors coordinates, while the decoder term protects refittable readouts weighted by the declared task prior.
\par}

\subsection{Scope of the guarantee}

{\color{black}
All formal claims above concern optimal regularized readouts refitted separately to the original and reconstructed representations. They do not imply that the original model's frozen downstream weights can consume the reconstruction, nor that labeled concepts become concentrated into a few sparse coordinates; Section~\ref{sec:boundary-tests} evaluates those questions separately. We also avoid describing the objective as mutual-information maximization or as a Fisher-information metric.
\par}

%% file: sections/experiments.tex
\section{\textcolor{black}{Experimental Setup}}
\label{sec:experiments}

{\color{black}
\paragraph{Controlled sparse generators.}
The controlled experiments use paired tied signed-TopK dictionaries on overcomplete, nonorthogonal sparse generators. The structured setting places lower-amplitude features in a declared protected subspace and evaluates unseen coefficient combinations from that same subspace. Comparisons share initialization, minibatches, architecture, sparsity, optimizer, and stopping rule. They include MSE, isotropic DPSAE, task-prior DPSAE, a frozen-task weighted-MSE loss, and a row-permuted task-prior control.

\paragraph{Language-model training and selection.}
The main language-model experiment trains 16,384-latent BatchTopK SAEs with target sparsity $k=32$ on normalized GPT-2-small residual-stream activations after block 8 \citep{radford2019language,bussmann2024batchtopk}. A disjoint 25M-token sweep selects the smallest decoder-loss weight that reduces exact decoder distortion by at least 10\% while keeping NMSE within 1\% of paired MSE. We freeze the selected $\gamma=0.03125$ before training three paired 100M-token runs and evaluating them on a reserved activation range. MSE and DPSAE pairs share initialization, token order, architecture, and optimization.

\paragraph{Metrics, uncertainty, and scope.}
The primary language-model metrics are exact held-out finite-group decoder distortion, activation NMSE, and inference $L_0$. Exact evaluation uses 16,384 tokens, groups of 128, identity targets, and 10,000 paired group-bootstrap resamples. Intervals quantify evaluation uncertainty conditional on each trained pair; three paired initializations test optimization repeatability rather than population generality. The frozen-network and Pythia concept evaluations use separately specified held-out data and are interpreted as boundary tests, not consequences of the decoder objective. Appendix~\ref{app:experimental-details} gives complete generators, splits, grids, evaluation geometry, controls, and provenance.
\par}

%% file: sections/results.tex
\section{\textcolor{black}{Experiments and Results}}
\label{sec:results}

\subsection{\textcolor{black}{A structured prior protects its declared task family}}

{\color{black}
The structured sparse generator tests whether a task covariance can direct capacity toward a lower-amplitude protected subspace. Held-out coefficients and examples are new, but their targets remain within the subspace that defines the prior. Task-prior DPSAE reduces median protected-task distortion from 0.1755 to 0.1332, a 25.3\% reduction, and has lower distortion than paired MSE in all ten seeds. Median NMSE changes from 0.0693 to 0.0667. A one-sided paired Wilcoxon test does not establish an NMSE improvement ($p=0.097$), so the protected-task gain has no observed reconstruction penalty in this generator. Relative to isotropic DPSAE, the frozen-task loss, and the row-permuted-prior control, task-prior DPSAE reduces protected-task distortion by 18.8\%, 16.4\%, and 14.1\%, respectively.
\par}

\begin{figure}[!htbp]
  \centering
  \includegraphics{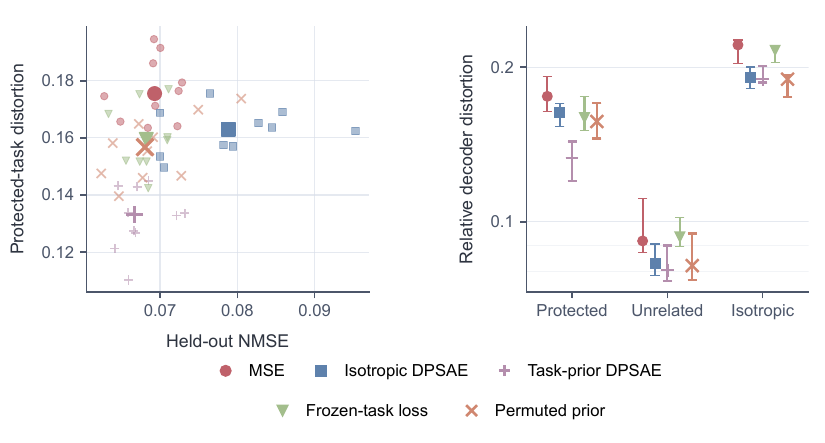}
  {\color{black}\caption{\textbf{Task-prior decoder preservation protects held-out combinations from the declared task subspace.} Left: individual paired seeds and medians for protected-task distortion versus reconstruction NMSE. Right: task-family distortions, with horizontally offset points showing medians and vertical whiskers showing the 10th--90th percentiles over ten seeds. Held-out coefficients and examples are fresh, but the targets remain within the protected latent span used to define the training prior.}\label{fig:structured-task-protection}}
\end{figure}

{\color{black}
The protected-task advantage persists across a 16-fold sweep of the prior weight: median reductions relative to paired MSE range from 23.9\% to 26.6\%, and every 10th-percentile seed reduction remains positive. The sparse generator does not reproduce the crossover from the separate commuting rank construction, so the sweep establishes robustness to task weight rather than a sparse phase transition. Appendix~\ref{app:structured-prior-sweep} gives the full sweep, and Appendix~\ref{app:isotropic-sparse} reports the secondary isotropic sparse control.
\par}

\subsection{\textcolor{black}{DPSAE preserves more GPT-2 readouts at matched reconstruction quality}}

\begin{figure}[!htbp]
  \centering
  \includegraphics{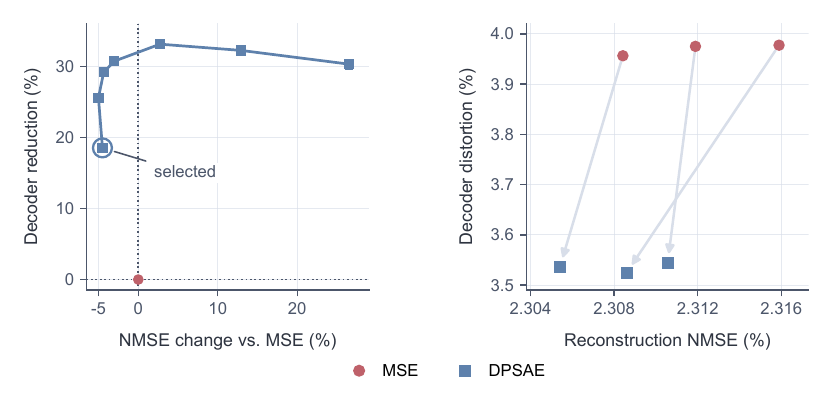}
  {\color{black}\caption{\textbf{DPSAE reduces held-out decoder distortion at matched reconstruction quality.} Left: one paired 25M-token run per weight on disjoint activations; the open ring marks the smallest weight that clears the predeclared reconstruction and decoder-distortion thresholds. Right: arrows connect three paired 100M-token MSE and DPSAE runs evaluated on a reserved activation range. Intervals resample held-out geometry groups and are conditional on the trained checkpoints.}\label{fig:lm-confirmation}}
\end{figure}

{\color{black}
The disjoint selection sweep traces the trade-off induced by the decoder weight (Figure~\ref{fig:lm-confirmation}, left). The predeclared rule selects $\gamma=0.03125$, the smallest weight that reduces exact decoder distortion by at least 10\% while keeping NMSE within 1\% of paired MSE, before the three paired 100M-token runs are trained. The sweep uses shorter 25M-token models and a separate activation range, so its points are not pooled with the final evaluation.

At the selected weight, DPSAE has lower decoder distortion and lower NMSE than paired MSE in all three runs (Figure~\ref{fig:lm-confirmation}, right). Exact held-out decoder-distortion reductions are 10.61--11.38\%, with every conditional 95\% interval excluding zero. NMSE changes by $-0.31$ to $-0.06\%$. Thus the tested operating point changes readout fidelity without exchanging it for worse reconstruction.
\par}

{\color{black}
Inference sparsity remains close to the $k=32$ training budget: MSE has $L_0=30.95$--$31.27$, and DPSAE has $L_0=30.51$--$30.80$. A tokenwise-TopK control, which retains exactly 32 activations for every token, reduces decoder distortion by 11.50\% (95\% CI 11.11--11.90\%) at a 0.81\% NMSE cost. This single 25M-token pair supports the narrower conclusion that the effect is not unique to BatchTopK's global support competition.

The sign of the reduction persists across the tested ridge strengths, group sizes, and grouping schemes. On cached activations, the decoder objective adds 21.3\% to the isolated SAE optimization step and about 6 MiB of peak GPU allocation. Appendix~\ref{app:lm-robustness} reports these robustness and compute measurements.
\par}

\subsection{\textcolor{black}{The readout gain is broad but not uniform}}

\begin{figure}[!htbp]
  \centering
  \includegraphics{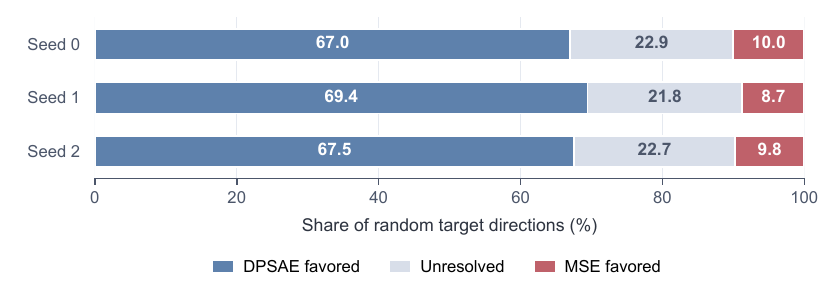}
  {\color{black}\caption{\textbf{The average readout gain is broad but not uniform.} Stacked bars show the share of 4,096 shared random target directions per held-out group that materially favor DPSAE, favor MSE, or remain unresolved for each paired run. A direction is material when its absolute advantage exceeds 5\% of that group's baseline mean task error. The directions live in finite sample space and should not be interpreted as semantic concepts.}\label{fig:taskwise-shares}}
\end{figure}

{\color{black}
Average decoder distortion hides how the gain is distributed across tasks. Every held-out group in all three paired runs has positive trace advantage, but every group also has material positive and negative eigenvalues; none of the advantage operators is positive semidefinite. Among shared random unit targets, 67.0--69.4\% materially favor DPSAE, 8.7--10.0\% favor MSE, and 21.8--22.9\% are unresolved at the declared 5\% scale (Figure~\ref{fig:taskwise-shares}). The improvement therefore spans many directions without providing uniform taskwise dominance. These finite-group directions are not semantic tasks, and Appendix~\ref{app:taskwise-audit} reports their weak cross-seed recurrence.
\par}

{\color{black}
The tested static covariance metrics do not reproduce the DPSAE reduction, but none reaches the matched NMSE regime, so this comparison remains unresolved. Appendix~\ref{app:static-controls} gives the separately selected high-weight controls and the matched-quality feasibility screen.
\par}

\subsection{\textcolor{black}{Readout fidelity does not imply cleaner concepts or uniform frozen behavior}}
\label{sec:boundary-tests}

\begin{figure}[!htbp]
  \centering
  \includegraphics{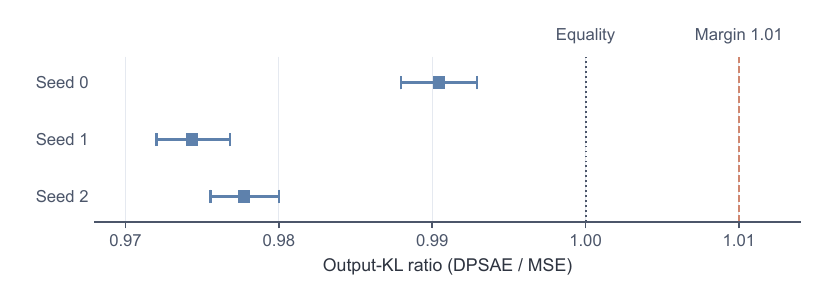}
  {\color{black}\caption{\textbf{All three GPT-2 pairs pass natural-text output-KL noninferiority.} Points show $\mathrm{KL}(p_{\mathrm{orig}}\Vert p_{\mathrm{DPSAE}})/\mathrm{KL}(p_{\mathrm{orig}}\Vert p_{\mathrm{MSE}})$ on 2,048 fresh FineWeb sequences. Whiskers are sequence-level paired-bootstrap 95\% intervals conditional on each trained pair; vertical lines mark equality and the predeclared 1.01 margin.}\label{fig:frozen-network-noninferiority}}
\end{figure}

{\color{black}
Refitted readout fidelity does not guarantee that the original downstream network will process the reconstruction in the same way. We insert each selected GPT-2 pair at block 8 of the frozen model and evaluate 2,048 fresh FineWeb sequences. The DPSAE/MSE output-KL ratios are 0.9905, 0.9744, and 0.9777, with 95\% intervals $[0.9880,0.9929]$, $[0.9720,0.9768]$, and $[0.9755,0.9800]$. Every upper endpoint lies below the predeclared 1.01 noninferiority margin (Figure~\ref{fig:frozen-network-noninferiority}). The intervals also lie below equality, but lower output KL remains secondary because the experiment was designed to test noninferiority.
\par}

{\color{black}
This compatibility is limited to the average natural-text output metric. On 2,048 separately specified IOI prompts, DPSAE reconstruction has 0.88--2.39 percentage points lower accuracy than paired MSE reconstruction, with every paired-bootstrap interval below zero. Its error in the correct-minus-incorrect name logit difference is also higher in all three pairs. Natural-text output-KL compatibility therefore does not imply uniform preservation of frozen task behavior.

Nor does improved average readout geometry imply that benchmark concepts concentrate into a few sparse coordinates. In one matched Pythia-160M block-8 pair, probes restricted to $k=5$ selected features have a family-macro DPSAE-minus-MSE AUROC difference of $-0.00387$ across 60 concept families, with a 95\% family-block interval of $[-0.00691,-0.00115]$. All ten probe-seed aggregates are negative. Because the trained checkpoint pair is the replication unit, this is a conditional result rather than evidence for a population-average disadvantage. It nevertheless rules out the stronger interpretation that the measured readout gain automatically improves concentration into one, two, or five benchmark features. Appendix~\ref{app:frozen-network} and Appendix~\ref{app:concept-pilot} give the complete boundary evaluations.
\par}

%% file: sections/related_work.tex
\section{Related Work}
\label{sec:related-work}

\paragraph{Prediction-based representation comparison.}
GULP defines representation distances through regularized linear prediction tasks and controls differences in predictive performance across those tasks \citep{boixadsera2022gulp}. Harvey et al.\ show that CKA, CCA, and related measures can be interpreted as average agreement between optimal linear readouts over distributions of decoding tasks \citep{harvey2024representational}. These works provide the prediction-operator geometry used here, but compare representations after they have been formed. Similarity-preserving distillation likewise uses representational relationships to train a compact student \citep{tung2019similarity}. DPSAE instead uses prediction-operator disagreement to train a sparse reconstruction and studies the taskwise freedom left by fixed MSE and sparsity.

\paragraph{SAE objectives and inductive biases.}
Language-model SAEs usually balance activation reconstruction and sparsity, with recent methods changing the support mechanism or encoder assumptions \citep{bricken2023monosemanticity,gao2025scaling,rajamanoharan2024jumping,bussmann2024batchtopk}. MP-SAE begins from hierarchical and nonlinear representational structure and introduces a residual-guided encoder suited to that structure \citep{costa2025matching}. Related work shows more generally that an SAE's assumptions determine which concepts it exposes \citep{hindupur2025projecting}. Temporal SAE adds a contrastive prior over adjacent tokens \citep{bhalla2026temporal}, while Aligned SAE uses a cross-modal energy prior that changes the learned dictionary without degrading reconstruction \citep{dhimoila2026crossmodal}. DPSAE studies a different ambiguity under a fixed sparse architecture: equal MSE and $L_0$ can induce different ridge-readout distortion profiles, and a target covariance weights that profile.

\paragraph{\textcolor{black}{Compression and model selection.}}
{\color{black}
Ayonrinde et al.\ cast SAE selection as a two-part coding problem, preferring the shortest quantized activation-and-decoder description at a fixed fidelity tolerance \citep{ayonrinde2024interpretability}. This addresses the rate and model-selection side of sparse compression. DPSAE instead fixes width and sparsity while changing the task-indexed distortion, without estimating operational description length or claiming MDL optimality.
\par}

\paragraph{Task-aware sparse representation learning.}
Supervised dictionary learning has long optimized sparse representations for specified classification or regression tasks rather than reconstruction alone \citep{mairal2012taskdriven}. Recent LLM methods bring related supervision into SAEs. Guided SAE binds labeled concepts to designated latent features \citep{harle2025guiding}; ClassifSAE jointly trains a classifier to concentrate task-relevant information in a sparse subspace \citep{lebail2026classifsae}; and AdaptiveK uses a supervised complexity probe to allocate input-dependent sparsity \citep{yao2026adaptivek}. These methods target specified labeled quantities. DPSAE instead weights a distribution of readouts that are refitted separately to the original and reconstructed activations. Its isotropic language-model objective requires no task labels.

\paragraph{Frozen-model objectives.}
End-to-end sparse dictionary learning adds output-distribution KL to activation reconstruction so that an SAE preserves the behavior of the frozen downstream model \citep{braun2024endtoend}. A short KL-and-MSE fine-tune can recover much of this benefit at lower cost \citep{karvonen2025revisiting}. These objectives preserve one realized downstream computation. \rev{DPSAE makes no frozen-network guarantee, so we test compatibility empirically rather than treating it as a consequence of the objective;} it preserves a specified distribution of refittable regularized linear readouts. To our knowledge, prior SAE objectives have not directly penalized disagreement between the regularized prediction operators induced by original and reconstructed activations.

%% file: sections/conclusion.tex



\section{Discussion, Limitations, and Future Work}
\label{sec:conclusion}

These results support a narrow but useful view of sparse compression: reconstruction error and sparsity leave substantial freedom in which refittable readouts survive, which DPSAE makes explicit by assigning a cost to disagreement between regularized prediction operators. A structured prior can steer capacity toward its declared task family in a controlled sparse generator, while the isotropic objective reduces held-out finite-group readout distortion by 10.6--11.4\% across three paired GPT-2 runs at matched reconstruction quality. This improvement is broad across sampled directions but is not uniform, and its meaning should remain distinct from other desirable properties of an SAE. The same GPT-2 reconstructions satisfy natural-text output-KL noninferiority, yet they perform worse on IOI, and one matched Pythia pair shows no improvement in concentration of benchmark concepts into a few sparse features. Refittable readout fidelity, compatibility with a frozen downstream computation, and sparse concept recovery are therefore separate empirical properties rather than interchangeable notions of representation quality.

The evidence also leaves important limits unresolved. The main language-model result concerns one layer of GPT-2 small and three paired initializations, so its intervals quantify evaluation uncertainty conditional on those trained checkpoints rather than population-level generality. The structured-prior result uses a synthetic task family, the taskwise audit studies finite-sample directions rather than semantic tasks, the static covariance controls never reach a matched reconstruction regime, and the Pythia concept result is conditional on one checkpoint pair. Future work will replicate the readout effect across layers, model families, widths, and sparse architectures; construct task priors from independently specified real prediction families; and compare adaptive decoder preservation with static baselines along genuinely matched reconstruction--sparsity frontiers. It would also be useful to test nonlinear and frozen readouts, connect readout preservation to causal interventions and feature semantics, and measure an operational rate that includes sparse supports, amplitudes, and dictionary cost. That broader rate--distortion view could determine when the additional readout fidelity is worth the representation complexity required to preserve it.

%% file: appendices/experimental_details.tex
\section{Experimental Details}
\label{app:experimental-details}

\subsection{\textcolor{black}{Finite-probe estimator and corpus aggregation}}
\label{app:estimator-details}

{\color{black}
For geometry groups $\mathcal G$, write $D_g^2=D_{\Sigma_{y,g}}^2(X_g,\Xhat_g)$ and $S_g=S_{\Sigma_{y,g}}(X_g)$. Assuming $\sum_gS_g>0$ for the first expression and $S_g>0$ for every group in the second, the two exact corpus summaries are
\begin{equation}
  R_{\mathcal G}^{\mathrm{sum}}
  =\frac{\sum_{g\in\mathcal G}D_g^2}{\sum_{g\in\mathcal G}S_g},
  \qquad
  R_{\mathcal G}^{\mathrm{mean}}
  =\frac1{|\mathcal G|}\sum_{g\in\mathcal G}\frac{D_g^2}{S_g}.
  \label{eq:corpus-aggregations}
\end{equation}
For sampled group statistics $\widehat D_{g,m}^2=\lVert\Delta_{X_g}(\Xhat_g)Y_{g,m}\rVert_F^2$ and $\widehat S_{g,m}=\lVert K_{X_g}Y_{g,m}\rVert_F^2$, the corresponding estimators are
\begin{equation}
  \widehat R_{\mathcal G,m}^{\mathrm{sum}}
  =\frac{\sum_g\widehat D_{g,m}^2}{\max\{\sum_g\widehat S_{g,m},\epsilon\}},
  \qquad
  \widehat R_{\mathcal G,m}^{\mathrm{mean}}
  =\frac1{|\mathcal G|}\sum_g
  \frac{\widehat D_{g,m}^2}{\max\{\widehat S_{g,m},\epsilon\}}.
  \label{eq:sampled-corpus-summaries}
\end{equation}
Language-model training and headline evaluation use the ratio of sums; the controlled synthetic experiments average groupwise ratios. Identity targets compute the isotropic trace exactly for a fixed group, not for a population over activations.

The language-model implementation draws $g_j\sim\mathcal N(0,I_n)$ and uses $y_j=\sqrt n\,g_j/\lVert g_j\rVert_2$. It omits the analytic $m^{-1/2}$ scale, which cancels in the ratio whenever the denominator clamp is inactive; no clamp activation occurs in the reported audit. The normalized numerator is unbiased for $D_I^2$, but the finite-$m$ self-normalized ratio is not exactly unbiased. Appendix~\ref{app:theory-proofs} gives the fixed-radius trace-estimator variance.

For the isotropic ratio-of-sums comparison, let $R_M=R_{\mathcal G}^{\mathrm{sum}}(X,\Xhat_M)$ and $R_D=R_{\mathcal G}^{\mathrm{sum}}(X,\Xhat_D)$. Proposition~\ref{prop:advantage-spectrum} gives
\begin{equation}
  R_M-R_D
  =\frac{\sum_g\tr Q_g}{\sum_g\lVert K_{X_g}\rVert_F^2}.
  \label{eq:trace-reconciliation}
\end{equation}
We calibrate ridge through the effective decoder degrees of freedom $\tr(K_X)/n$, using a target of $0.25$ in the language-model experiments. Batched FP32 Cholesky solves apply the operators without materializing a task covariance.
\par}

\subsection{\textcolor{black}{Language-model protocol and paired GPT-2 evaluation}}
\label{app:lm-protocol}

\paragraph{Language-model representation and preprocessing.}
{\color{black}The main GPT-2 experiment uses \texttt{openai-community/gpt2}, with 768-dimensional residual-stream activations after block 8. A fixed training-calibration mean vector and one global RMS scalar normalize all later activations. Geometry groups are not centered independently. We calibrate the ridge penalty by a target effective degrees-of-freedom fraction, $\tr(K_X)/n$. The untied nonnegative BatchTopK SAE has 16,384 latents, a learned decoder bias, unit-normalized decoder columns, ReLU preactivations, and a global training budget of $k=32$ activations per token on average. SAE encoding and decoding use BF16 autocast on CUDA, while reconstructions passed to the decoder objective and all ridge solves use FP32.}

\paragraph{\textcolor{black}{Data splits and model selection.}}
{\color{black}
Decoder-weight and architecture choices use the designated 180M--185M FineWeb range \citep{penedo2024fineweb}. The clean readout evaluation uses the disjoint 190M--195M range. A frozen-network diagnostic opened 195M--200M once and did not enter model or hyperparameter selection. After the output-KL metric, 1.01 margin, sequence sampler, sample size, bootstrap, and secondary metrics were fixed, the frozen-network evaluation opened the fresh 200M--210M range. Exact readout evaluation uses 16,384 tokens, groups of 128, calibrated ridge, identity targets, and 10,000 paired group-bootstrap resamples.
\par}

\paragraph{\textcolor{black}{Paired GPT-2 confirmation.}}
{\color{black}The selected $\gamma=0.03125$ is frozen before training the three paired runs. Each model sees 100,001,792 training tokens. Initialization seeds are 0, 1, and 2, while token order and fixed-radius isotropic probe streams are matched within each pair. Exact evaluation uses ridge $\lambda=1.6049035191535947$ and reports the ratio-of-sums distortion $\sum_g\lVert K(X_g)-K(\widehat X_g)\rVert_F^2/\sum_g\lVert K(X_g)\rVert_F^2$. Seedwise reductions and conditional 95\% intervals are 10.61\% [10.15, 11.07], 11.38\% [10.95, 11.82], and 10.84\% [10.28, 11.40]. The complete run uses clean revision \texttt{79607d4}; its run manifest records code, configuration, cache, checkpoint, and evaluation hashes.}

\paragraph{Decoder-weight selection sweep.}
{\color{black}The selection sweep trains paired 25M-token models at $\gamma\in\{0.03125,0.0625,0.09375,0.125,0.25,0.5,1\}$ with matched initialization, token order, probe stream, optimizer, and $k=32$ budget. The predeclared rule chooses the smallest weight with at least 10\% lower exact decoder distortion and NMSE no greater than 1.01 times MSE. This selects $\gamma=0.03125$ for the paired 100M-token evaluation; the full sweep remains a selection-set trade-off curve.}

\subsection{\textcolor{black}{Static covariance controls}}
\label{app:static-controls}

\begin{figure}[!htbp]
  \centering
  \includegraphics{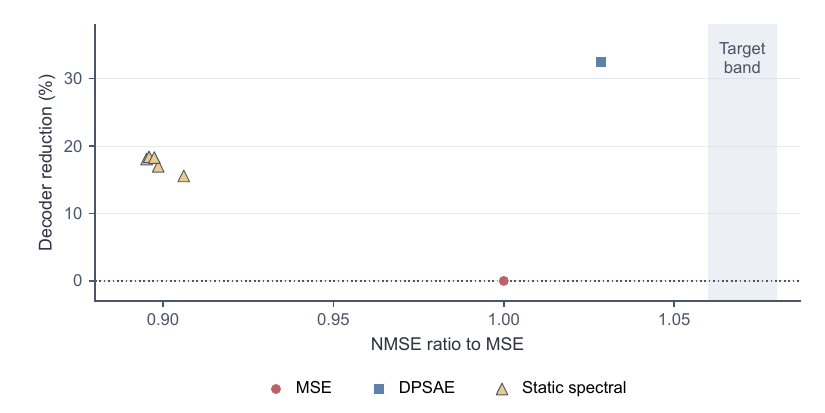}
  {\color{black}\caption{\textbf{The tested static spectral coefficients do not reach the matched-NMSE target.} The shaded band marks the predeclared $[1.06,1.08]$ DPSAE/MSE NMSE-ratio range. All five static candidates fall outside it, so the screen cannot compare a static metric with DPSAE at matched reconstruction quality.}\label{fig:static-feasibility}}
\end{figure}

\begin{figure}[!htbp]
  \centering
  \includegraphics{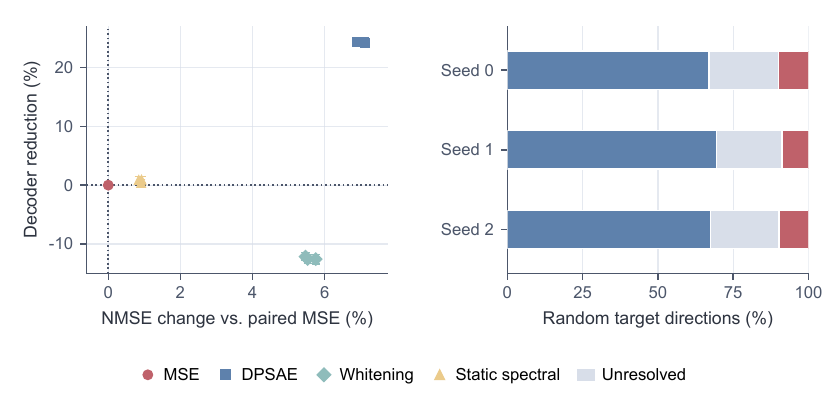}
  {\color{black}\caption{\textbf{Static covariance controls remain unmatched, while the average DPSAE advantage is taskwise mixed.} Left: separately selected high-weight DPSAE, whitening, and static spectral points; these controls do not share an NMSE operating point. Right: random-direction shares from the three paired GPT-2 runs. The main text uses the enlarged taskwise panel in Figure~\ref{fig:taskwise-shares}.}\label{fig:static-and-taskwise-audit}}
\end{figure}

{\color{black}
For calibration covariance $C$, the whitening operator is a floored inverse covariance square root. The static spectral operator is the square root of $M_\lambda=C(C+\lambda I)^{-2}$, so its residual norm implements the rank-relaxation weighting. Each coefficient is selected from $\{0.125,0.25,0.5,1\}$ under an NMSE cap on fresh activations, then evaluated across three paired 100M-token runs on 185M--190M. These control runs use revision \texttt{38b4c3b}; the Exp08 manifest pins their result artifact and configuration hash.

At $\gamma=0.25$, DPSAE reduces decoder distortion by 24.23--24.39\% at a 6.91--7.11\% NMSE cost. The selected static spectral loss changes distortion by 0.37--0.91\% at a 0.85--0.92\% NMSE cost, while selected whitening worsens distortion by 12.16--12.57\%. These points were selected separately and are not matched in NMSE, so they do not identify a like-for-like effect comparison (Figure~\ref{fig:static-and-taskwise-audit}, left).

The later matched-quality screen is a separate preregistered feasibility test. Its $[1.06,1.08]$ band targets the roughly 7\% NMSE cost of the earlier high-weight DPSAE runs. A co-trained 25M-token MSE/DPSAE anchor has an NMSE ratio of 1.0285 and a 32.51\% decoder-distortion reduction. The five static spectral coefficients $\beta\in\{2,4,8,16,32\}$ produce NMSE ratios 0.9061, 0.8986, 0.8952, 0.8959, and 0.8974, all outside the specified band (Figure~\ref{fig:static-feasibility}). None reaches the regime needed for a matched comparison, leaving the static-baseline question unresolved.
\par}

\noindent\begin{minipage}{\linewidth}
\subsection{\textcolor{black}{Taskwise advantage audit}}
\label{app:taskwise-audit}

{\color{black}
The exact spectrum audit reuses the 16,384 held-out tokens, 128 groups of 128, and the calibrated ridge. For each group we form $Q_g$ in float64, check symmetry and trace reconciliation, and evaluate 4,096 random unit directions drawn once and reused across runs. Materiality is an absolute task advantage above 5\% of the group's baseline mean task error. A shared direction materially favors DPSAE in all three runs with probability 32.73\%, versus 31.78\% under conditional independence; pairwise score correlations are 0.054--0.075. These weak recurrence statistics do not identify a stable semantic subspace. The source summary is \path{artifacts/exp08_experiment_figure/task_spectrum/advantage_spectrum_summary.json}; compact shares are versioned in \path{data/fig_lm_task_shares.csv}.
\par}

\end{minipage}

{\color{black}\subsection{Frozen-network compatibility}\label{app:frozen-network}}

{\color{black}
The evaluation inserts each of the three selected 100M-token GPT-2 BatchTopK pairs at the original block-8 residual-stream site and leaves every downstream parameter fixed. It samples 2,048 nonoverlapping length-256 sequences by a deterministic permutation of aligned blocks from FineWeb positions 200M--210M. The primary metric is the ratio $\sum_t\mathrm{KL}(p_{\mathrm{orig},t}\Vert p_{\mathrm{DPSAE},t})/\sum_t\mathrm{KL}(p_{\mathrm{orig},t}\Vert p_{\mathrm{MSE},t})$ over evaluated tokens. Sequence-level paired bootstraps use 10,000 draws, and every run must have an upper 95\% endpoint below 1.01. An identity hook reproduces all logits exactly; the mean-ablation control replaces the normalized activation with zero before denormalization.
\par}

{\color{black}
All three natural-text pairs meet the noninferiority margin, with slightly better activation NMSE for DPSAE and effectively matched inference sparsity (Table~\ref{tab:frozen-boundaries}). The inference-$L_0$ differences are $+0.0082$, $+0.0010$, and $-0.0106$ activations for seeds 0--2. Cross-entropy, top-1 agreement, and next-token accuracy are directionally favorable or unresolved on natural text, but they remain secondary to the frozen KL ratio.
\par}

{\color{black}
The controlled IOI evaluation uses 2,048 held-out prompts generated before evaluation. The original model is correct on 99.66\% of them. DPSAE reconstruction is less accurate than paired MSE reconstruction in every run, and its absolute error in the correct-minus-incorrect name logit difference is higher by 0.1206, 0.0759, and 0.1085, with all prompt-bootstrap intervals above zero. The result shows why average natural-text output-KL compatibility should not be restated as uniform frozen-behavior preservation.
\par}

{\color{black}
\begin{table}[H]
  \centering
  \color{black}
  \scriptsize
  {\color{black}\caption{\textbf{Natural-text output-KL noninferiority passes, while the secondary IOI accuracy endpoint worsens.} The output-KL ratio is $\mathrm{KL}(p_{\mathrm{orig}}\Vert p_{\mathrm{DPSAE}})/\mathrm{KL}(p_{\mathrm{orig}}\Vert p_{\mathrm{MSE}})$; the IOI accuracy difference is DPSAE minus MSE. Intervals resample natural sequences or IOI prompts and are conditional on each trained checkpoint pair.}\label{tab:frozen-boundaries}}
  \begin{tabular}{crrr}
    \toprule
    Seed & Natural KL ratio [95\% CI] & Activation NMSE ratio & IOI $\Delta$accuracy (pp) [95\% CI] \\
    \midrule
    0 & 0.9905 [0.9880, 0.9929] & 0.9975 & $-2.39$ [$-3.22$, $-1.61$] \\
    1 & 0.9744 [0.9720, 0.9768] & 0.9967 & $-0.88$ [$-1.56$, $-0.20$] \\
    2 & 0.9777 [0.9755, 0.9800] & 0.9988 & $-1.51$ [$-2.20$, $-0.88$] \\
    \bottomrule
  \end{tabular}
\end{table}
}

{\color{black}\subsection{Standard-concept concentration pilot}\label{app:concept-pilot}}

{\color{black}
The evaluation reuses one 25M-token Pythia-160M-deduped block-8 BatchTopK pair with width 16,384, target $L_0=32$, and decoder weight $\gamma=0.03125$. Its evaluation-split DPSAE/MSE NMSE ratio is 1.0031; the methods' relative $L_0$ errors are 0.00117 and 0.00144, and their paired relative $L_0$ difference is 0.00027. The pair is therefore matched closely enough in reconstruction and sparsity to test concept concentration.
\par}

{\color{black}
The benchmark contains 113 frozen labeled tasks grouped into 60 concept families. Sparse probes use the pinned \texttt{sae-probes} L1 procedure at $k\in\{1,2,5\}$ and ten probe seeds. Companion L2 logistic probes evaluate the original residual stream, each full reconstruction, and each full sparse code. The primary analysis first averages probe seeds within each task, then computes the family-macro DPSAE-minus-MSE test-AUROC difference at $k=5$ and resamples the 60 family blocks 10,000 times. Missing or extra task--$k$--seed--method cells fail the run rather than being dropped.
\par}

\begin{figure}[!htbp]
  \centering
  \includegraphics{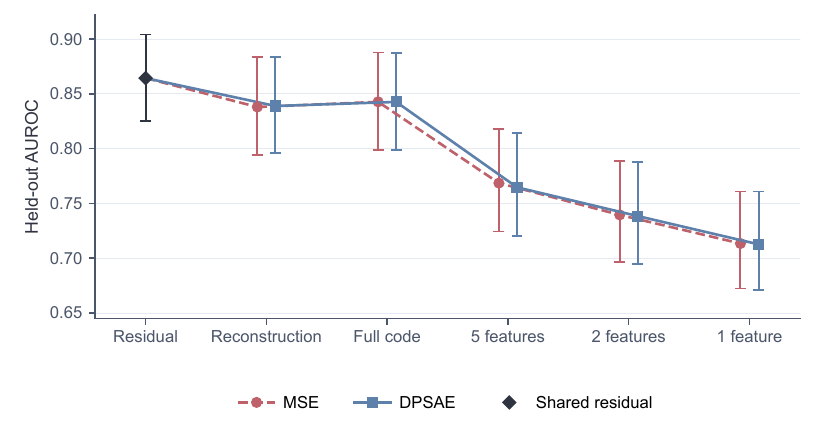}
  {\color{black}\caption{\textbf{One matched Pythia pair does not improve standard sparse-concept probes.} The representation ladder separates original-activation recoverability, full-reconstruction and full-code preservation, and concentration into $k=5,2,1$ selected features. Intervals resample the fixed concept-family blocks after averaging the ten probe seeds within each task. Because no additional training pairs were evaluated, the result is conditional on this checkpoint pair.}\label{fig:concept-ladder}}
\end{figure}

{\color{black}
The primary difference is $-0.003865$, with a 95\% family-block interval of $[-0.006910,-0.001150]$. All ten probe-seed macro differences are negative, from $-0.006643$ to $-0.000893$, while 29 of 60 individual family means are positive. Full-reconstruction and full-code differences are near zero, and the sparse representation ladder in Figure~\ref{fig:concept-ladder} shows no DPSAE concentration gain at $k=5,2,$ or $1$. Because the preregistered primary aggregate did not improve, we did not run additional training seeds, context mining, or API feature labeling.
\par}

\subsection{Evaluation geometry and computational cost}
\label{app:lm-robustness}

\paragraph{Geometry audit.}
\textcolor{black}{The audit reuses the three paired GPT-2 checkpoints and changes one evaluation choice at a time.} Ridge penalties target effective-DoF fractions $0.125$, $0.25$, and $0.5$; group-size checks use 64, 128, and 256 tokens; grouping checks compare contiguous, globally shuffled, and document-balanced groups at size 128. Every seed retains a positive paired reduction in every condition. Because group size changes the finite-group operator itself, the resulting magnitudes should not be compared as measurements of one fixed quantity.

\begin{figure}[!htbp]
  \centering
  \includegraphics{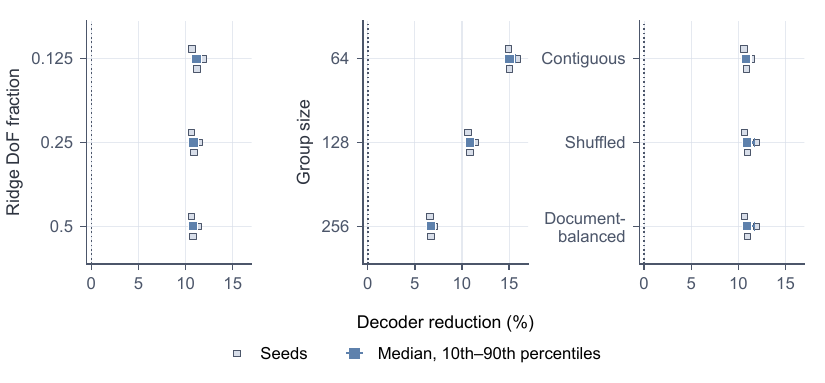}
  {\color{black}\caption{\textbf{The sign of the DPSAE advantage survives the tested evaluation geometries.} Points show paired reductions and bars show the median with 10th--90th percentiles. Left: ridge effective-DoF fraction. Center: geometry-group size. Right: contiguous, shuffled, and document-balanced grouping. These are evaluation-only checks on fixed checkpoints; changing group size or ridge changes the measured operator.}
  \label{fig:lm-robustness}}
\end{figure}

\paragraph{Computational cost.}
The isolated optimization benchmark uses cached normalized activation batches, so it excludes the shared language-model forward pass. Median step time is 6.494 ms for MSE and 7.878 ms for DPSAE, a 21.3\% increase. Corresponding throughput is 315,385 and 259,964 activation tokens per second. Peak allocated memory is 1.3422 and 1.3485 GiB, respectively. \textcolor{black}{The reported robustness and optimization suite uses 1.113 active single-GPU hours, including 1.075 hours of SAE training; queue idle time and figure rendering are excluded.}

\paragraph{Single-seed generality diagnostics.}
{\color{black}GPT-2 block 4 clears the 10\% decoder-reduction threshold in one paired diagnostic, while Pythia block 8 reaches a 9.13\% reduction. These screens use neither three paired runs nor the complete matched-quality protocol, so they do not establish cross-layer or cross-model generality.}

\subsection{\textcolor{black}{Isotropic sparse control}}
\label{app:isotropic-sparse}

{\color{black}
In the rank-24 numerical construction, truncated SVD matches the decoder-distortion tail in Theorem~\ref{thm:isotropic-rank-relaxation} at every retained rank, with maximum absolute error $2.22\times10^{-15}$. This verifies the implementation rather than adding empirical evidence for the theorem. Replacing the rank constraint with a TopK bottleneck on an overcomplete, nonorthogonal sparse generator changes the allocation: isotropic DPSAE lowers decoder distortion relative to paired MSE in all ten seeds, with a median reduction of 15.5\% and a range of 8.9--21.7\%, while median NMSE increases by 12.4\%. Thus sparsity permits departures from the rank-relaxed allocation, but this isotropic control does not produce a matched-reconstruction improvement or identify a privileged low-variance subspace.
\par}

\subsection{Structured-prior weight sweep}
\label{app:structured-prior-sweep}

The sparse generator uses the crossover from the commuting two-direction rank construction only to set a dimensionless weight scale. Across relative weights $0.25$ to $4$, median protected-task reductions remain between 23.9\% and 26.6\%, but neither a crossover nor a monotone response appears. The result shows that the protected-task advantage is stable across this range; it does not confirm that the sparse noncommuting system inherits the rank threshold.

\begin{figure}[!htbp]
  \centering
  \includegraphics{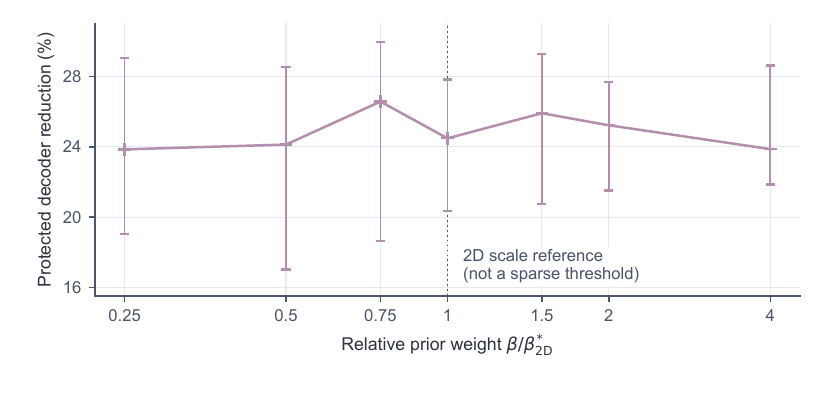}
  \caption{\textbf{Structured-prior protection is stable across task weights, without a sparse crossover.} Points show median protected-task distortion reduction relative to paired MSE over ten seeds; whiskers span the 10th--90th percentiles. The vertical reference marks the crossover scale from the separate commuting two-direction rank construction, not a predicted transition for this sparse generator.}
  \label{fig:structured-prior-weight-sweep}
\end{figure}

\subsection{Optimization and architecture audits}
\label{app:optimization-audits}

\paragraph{Finite-probe gradient audit.}
\label{app:gradient-audit}
For each of the six confirmation checkpoints, the audit compares gradients from 16-probe banks against identity-target reference gradients in both reconstruction and row-Gram coordinates. It uses 256 independent banks per batch and paired self-normalized and fixed-denominator estimates. The conservative columns below take the worse of the two coordinate systems. ``Expected UCB'' is the familywise paired confidence upper bound on sampled-minus-fixed gradient error; ``finite error'' and cosine describe the 256-bank mean rather than one 16-probe update.
\begin{table}[H]
  \centering
  \small
  {\color{black}\caption{\textbf{The finite-probe decoder gradient is expectation-aligned but noisy at finite sample size.} Across all checkpoints, the familywise upper confidence bound on expected gradient error remains below 0.53\%, whereas the 256-bank mean retains 13.8--14.1\% relative error.}}
  \label{tab:gradient-fidelity}
  \begin{tabular}{lrrrr}
    \toprule
    Checkpoint & Median UCB (\%) & P90 UCB (\%) & Finite error (\%) & \shortstack{256-bank\\mean cosine} \\
    \midrule
    MSE s0   & 0.470 & 0.528 & 14.12 & 0.9902 \\
    DPSAE s0 & 0.464 & 0.522 & 13.80 & 0.9906 \\
    MSE s1   & 0.470 & 0.529 & 14.14 & 0.9902 \\
    DPSAE s1 & 0.464 & 0.521 & 13.82 & 0.9906 \\
    MSE s2   & 0.470 & 0.528 & 14.12 & 0.9902 \\
    DPSAE s2 & 0.464 & 0.521 & 13.87 & 0.9905 \\
    \bottomrule
  \end{tabular}
\end{table}

The denominator coefficient of variation is 1.143\% at the median and 1.216\% at the 90th percentile, with no denominator or target-RMS clamp activations. Mean error follows the expected $m^{-1/2}$ Monte Carlo rate. This audit applies before the active-set Jacobian, MSE mixing, $\gamma$ scaling, gradient clipping, and Adam, so it does not certify the complete optimizer update.

\paragraph{Matched-NMSE and architecture screens.}
The architecture screen replaces BatchTopK with tokenwise TopK or attempts a gated JumpReLU extension. Tokenwise TopK keeps 32 preactivations per token and uses the selected decoder weight in a single paired 25M-token run.

\paragraph{JumpReLU validity screen.}
\label{app:jumprelu-screen}
The JumpReLU extension uses $z_j=a_j\mathbf{1}\{a_j>\theta_j\}$ and a shared penalty $(\lVert z\rVert_0/k-1)^2$. We parameterize thresholds in log space and train them with rectangular-kernel pseudo-gradients \citep{rajamanoharan2024jumping}. Scores and threshold gradients use FP32, and a one-time first-batch quantile sets the initial average $L_0=k$.

{\color{black}The final JumpReLU validity study trains the paired objectives for 25,001,984 tokens at shared sparsity-loss weights $2,4,8,$ and $16$. Selection uses only held-out $L_0$, late-window drift, finite values, threshold stability, dead-feature fraction, and run provenance; NMSE, decoder distortion, and language-model loss are not inspected. The respective DPSAE/MSE $L_0$ values are $37.73/38.55$, $37.38/36.93$, $36.10/36.11$, and $35.88/35.61$. Every setting falls outside the prespecified $[30.4,33.6]$ band, and maximum dead-feature fractions range from 48.5\% to 54.4\%, above the prespecified 10\% maximum. No weight is selected, so the JumpReLU comparison remains unidentified rather than providing evidence for or against DPSAE.}

\paragraph{\textcolor{black}{Reproducibility and provenance.}}
{\color{black}
The paired GPT-2 evaluation, gamma sweep, task spectrum, robustness audit, and structured-prior sweep use revision \texttt{79607d4}. Their versioned run manifest records code, configuration, input, checkpoint, and result hashes. The static-control runs use revision \texttt{38b4c3b}; the later matched-quality screen and frozen-network evaluation use revision \texttt{4fe3e91}. The gradient audit uses revision \texttt{46eb22a}, and the full-horizon JumpReLU study uses revision \texttt{6a2f338}.

The release is bound to core-manifest SHA-256 prefix \texttt{db9deb08d28c} and publication-payload SHA-256 prefix \texttt{15887dd617c6}. The versioned files \path{data/arxiv_closure_payload.json}, \path{data/arxiv_closure_payload_manifest.json}, \path{data/arxiv_closure_summary.csv}, and \path{data/arxiv_closure_figure_manifest.json} retain the full digests, experiment decisions, and exact figure hashes. The secondary record \path{data/exp09_frozen_network_secondary_summary.json} binds the natural-text and IOI values above to their full-result hashes. Generated checkpoints and full per-example artifacts remain outside the manuscript repository.
\par}

%% file: appendices/theory_proofs.tex
\section{Proofs for the Readout-Distortion Results}
\label{app:theory-proofs}

\paragraph{Proof of Proposition~\ref{prop:equal-error-separation}.}
Fix $\lambda>0$, let $\tau=2\lambda$, and choose $a,\delta>0$. Set
\[
  X=aI_2,
  \qquad
  W=I_2,
  \qquad
  Z_1=\begin{pmatrix}a+\delta&0\\0&a\end{pmatrix},
  \qquad
  Z_2=\begin{pmatrix}a&0\\0&a+\delta\end{pmatrix}.
\]
The reconstructions are $\Xhat_j=Z_jW=Z_j$. Each code has one nonzero entry per row, so $L_0(Z_1)=L_0(Z_2)=1$, and both reconstruction errors equal $\delta^2$. Define
\[
  \kappa(t)=\frac{t^2}{t^2+\tau},
  \qquad
  \eta=\kappa(a+\delta)-\kappa(a)>0.
\]
Direct substitution into Equation~\ref{eq:ridge-hat} gives
\[
  K_X=\kappa(a)I_2,
  \quad
  \mathcal A_X(\Xhat_1)=\begin{pmatrix}\eta^2&0\\0&0\end{pmatrix},
  \quad
  \mathcal A_X(\Xhat_2)=\begin{pmatrix}0&0\\0&\eta^2\end{pmatrix}.
\]
Their traces, and hence their isotropic decoder distances, agree. Their difference is $\operatorname{diag}(\eta^2,-\eta^2)$, which is indefinite. The first reconstruction has lower distortion on $e_2$, and the second has lower distortion on $e_1$. \hfill$\square$

\paragraph{Proof of Proposition~\ref{prop:advantage-spectrum}.}
Equation~\ref{eq:task-distortion} gives
\[
  d_{\Xhat_M}(y)-d_{\Xhat_D}(y)
  =y^\top\!\left(\mathcal A_X(\Xhat_M)-\mathcal A_X(\Xhat_D)\right)y
  =y^\top Qy.
\]
Rayleigh--Ritz gives the extrema over unit targets, and nonnegativity for every target is equivalent to $Q\succeq0$. Taking expectation under $\E[yy^\top]=\Sigma_y$ gives
\[
  \E[y^\top Qy]=\tr(\Sigma_yQ).
\]
If $Q$ is indefinite, unit eigenvectors associated with a positive and a negative eigenvalue define rank-one task priors that prefer opposite reconstructions. \hfill$\square$

\paragraph{Proof of Theorem~\ref{thm:isotropic-rank-relaxation}.}
Write $\tau=n\lambda$. The compact SVD of $X$ gives
\begin{equation}
  K_X
  =U\operatorname{diag}\!\left(
  \frac{\sigma_1^2}{\sigma_1^2+\tau},\ldots,
  \frac{\sigma_s^2}{\sigma_s^2+\tau}
  \right)U^\top
  =U\operatorname{diag}(q_1,\ldots,q_s)U^\top.
  \label{eq:ridge-spectral-form}
\end{equation}
For any $\Xhat$ with rank at most $r$, $K_{\Xhat}$ also has rank at most $r$. Eckart--Young therefore gives
\[
  \lVert K_X-K_{\Xhat}\rVert_F^2
  \geq
  \min_{\operatorname{rank}(M)\leq r}\lVert K_X-M\rVert_F^2
  =\sum_{i>r}q_i^2.
\]
The truncated SVD $X_r=U_r\operatorname{diag}(\sigma_1,\ldots,\sigma_r)V_r^\top$ satisfies
\[
  K_{X_r}=U_r\operatorname{diag}(q_1,\ldots,q_r)U_r^\top,
\]
so it attains this lower bound. \hfill$\square$

When $0<r<s$ and $q_r>q_{r+1}$, the best rank-$r$ approximation to $K_X$ is unique. Every optimizer then satisfies
\[
  K_{\Xhat}=U_r\operatorname{diag}(q_1,\ldots,q_r)U_r^\top,
  \qquad
  \Xhat\Xhat^\top=U_r\operatorname{diag}(\sigma_1^2,\ldots,\sigma_r^2)U_r^\top.
\]
With the same feature dimension, this is equivalent to $\Xhat=X_rQ$ for some orthogonal $Q$. At a tied cutoff, any corresponding subspace within the tied left-singular eigenspace is optimal.

\paragraph{Proof of Corollary~\ref{cor:structured-prior-selection}.}
Because $XX^\top$ and $\Sigma_y$ are symmetric and commute, they admit a simultaneous orthonormal eigenbasis. In this basis,
\[
  K_X\Sigma_y^{1/2}
  =U\operatorname{diag}(q_i\sqrt{\omega_i})U^\top.
\]
For any rank-$r$ reconstruction, $K_{\Xhat}\Sigma_y^{1/2}$ has rank at most $r$, while
\[
  D_{\Sigma_y}^2(X,\Xhat)
  =\lVert(K_X-K_{\Xhat})\Sigma_y^{1/2}\rVert_F^2.
\]
Eckart--Young lower-bounds this expression by the sum of the omitted squared singular values of $K_X\Sigma_y^{1/2}$, namely $\sum_{i\notin S_r}\omega_iq_i^2$. A reconstruction whose row Gram matrix is
\[
  \Xhat\Xhat^\top=\sum_{i\in S_r}\sigma_i^2u_iu_i^\top
\]
has ridge operator $\sum_{i\in S_r}q_iu_iu_i^\top$ and attains the bound. The argument does not require $\Sigma_y$ to be invertible. \hfill$\square$

\paragraph{Proof of Proposition~\ref{prop:zero-distance}.}
Let $G=XX^\top$ and $\tau=n\lambda>0$. The dual form of the ridge operator is
\[
  K_X=G(G+\tau I)^{-1}=I-\tau(G+\tau I)^{-1}.
\]
Equality of two ridge operators is therefore equivalent to equality of the corresponding resolvents, and hence to equality of their row Gram matrices. If $\Sigma_y\succ0$, then
\[
  D_{\Sigma_y}^2(X,\Xhat)
  =\lVert\Delta_X(\Xhat)\Sigma_y^{1/2}\rVert_F^2=0
\]
is equivalent to $\Delta_X(\Xhat)=0$. If $X,\Xhat\in\R^{n\times d}$ have equal row Grams, take thin singular-value decompositions with the same left singular vectors and singular values, then extend their right singular bases to orthogonal bases of $\R^d$. The resulting change of right basis gives $\Xhat=XQ$ for an orthogonal $Q$. \hfill$\square$

For singular $\Sigma_y$, zero weighted distance only requires $\Delta_X(\Xhat)\Sigma_y^{1/2}=0$. The objective can therefore ignore changes on the nullspace of its task prior.

\paragraph{Proof of the task-ellipsoid bound.}
For $y\in\operatorname{range}(\Sigma_y)$, write $y=\Sigma_y^{1/2}z$ with the minimum-norm choice $z=\Sigma_y^{\dagger/2}y$. Then
\[
  \lVert\Delta_X(\Xhat)y\rVert_2^2
  \leq
  \lVert\Delta_X(\Xhat)\Sigma_y^{1/2}\rVert_F^2\lVert z\rVert_2^2
  =D_{\Sigma_y}^2(X,\Xhat)y^\top\Sigma_y^\dagger y.
\]
Equivalently,
\[
  \sup_{\substack{y\in\operatorname{range}(\Sigma_y)\\y^\top\Sigma_y^\dagger y\leq1}}
  \lVert\Delta_X(\Xhat)y\rVert_2^2
  =\lVert\Delta_X(\Xhat)\Sigma_y^{1/2}\rVert_{\mathrm{op}}^2
  \leq D_{\Sigma_y}^2(X,\Xhat).
\]

\paragraph{MSE controls decoder distance under bounded norms.}
Let $G=XX^\top$ and $\widehat G=\Xhat\Xhat^\top$. The resolvent identity and $G+\tau I,\widehat G+\tau I\succeq\tau I$ give
\[
  \lVert K_X-K_{\Xhat}\rVert_F
  \leq\frac1\tau\lVert G-\widehat G\rVert_F
  \leq\frac{\lVert X\rVert_{\mathrm{op}}+\lVert\Xhat\rVert_{\mathrm{op}}}{\tau}
  \lVert X-\Xhat\rVert_F.
\]
Consequently,
\[
  D_{\Sigma_y}^2(X,\Xhat)
  \leq
  \frac{\lVert\Sigma_y\rVert_{\mathrm{op}}(\lVert X\rVert_{\mathrm{op}}+\lVert\Xhat\rVert_{\mathrm{op}})^2}{(n\lambda)^2}
  \lVert X-\Xhat\rVert_F^2.
\]
Thus norm-controlled Euclidean reconstruction bounds decoder distortion at fixed nonzero ridge. Proposition~\ref{prop:zero-distance} supplies the nonconverse.

\paragraph{Ridge nondegeneracy.}
If $X$ has full row rank and smallest singular value $\sigma_{\min}(X)>0$, Equation~\ref{eq:ridge-spectral-form} yields
\[
  \lVert I-K_X\rVert_{\mathrm{op}}
  =\frac{n\lambda}{\sigma_{\min}(X)^2+n\lambda}.
\]
Hence $K_X\to I$ as $\lambda\to0$. In this regime, any two fixed full-row-rank representations have nearly identical ridge operators, so calibrating $\tr(K_X)/n$ away from one prevents a degenerate decoder distance.

\paragraph{Fixed-radius trace-estimator variance.}
Let $s_j=\sqrt n\,g_j/\lVert g_j\rVert_2$ for independent $g_j\sim\mathcal N(0,I_n)$, let $C$ be fixed, and define
\[
  \widehat T_m=\frac1m\sum_{j=1}^m\lVert Cs_j\rVert_2^2,
  \qquad
  B=C^\top C.
\]
Since $s_j$ is uniform on the sphere of radius $\sqrt n$,
\[
  \E[s_{j,a}s_{j,b}s_{j,c}s_{j,d}]
  =\frac{n}{n+2}
  (\delta_{ab}\delta_{cd}+\delta_{ac}\delta_{bd}+\delta_{ad}\delta_{bc}).
\]
It follows that
\begin{align}
  \E[\widehat T_m]&=\tr B=\lVert C\rVert_F^2,\\
  \operatorname{Var}(\widehat T_m)
  &=\frac{2}{m(n+2)}
  \left[n\lVert B\rVert_F^2-(\tr B)^2\right].
  \label{eq:fixed-radius-variance}
\end{align}
For $C\neq0$, with $r_{\mathrm{eff}}(C)=\lVert C\rVert_F^4/\lVert C^\top C\rVert_F^2$, the relative standard deviation is
\[
  \frac{\operatorname{sd}(\widehat T_m)}{\lVert C\rVert_F^2}
  =\sqrt{\frac{2}{m(n+2)}\left(\frac{n}{r_{\mathrm{eff}}(C)}-1\right)}.
\]
Taking $C=\Delta_X(\Xhat)$ gives the isotropic numerator used by the language-model DPSAE before its target-RMS safety clamp, which did not activate in the reported audit. Taking $C=\Delta_X(\Xhat)\Sigma_y^{1/2}$ gives a transformed-sphere estimator, but it is not the concatenated structured-prior estimator used in the synthetic experiment. This result does not make the finite-probe ratio in Equation~\ref{eq:sampled-relative-loss} unbiased.

\paragraph{Finite-corpus trace reconciliation.}
For isotropic group $g$, let $D_{m,g}^2=\lVert\Delta_X(\Xhat_m)\rVert_F^2$ and $S_g=\lVert K_{X_g}\rVert_F^2$. Proposition~\ref{prop:advantage-spectrum} gives $\tr Q_g=D_{M,g}^2-D_{D,g}^2$. Summing and dividing by $\sum_gS_g$ proves Equation~\ref{eq:trace-reconciliation}. For a mean of groupwise ratios, the corresponding difference is
\[
  \frac1{|\mathcal G|}\sum_g\frac{\tr Q_g}{S_g},
\]
which generally differs from the ratio-of-sums identity.